\documentclass[10pt,twocolumn,letterpaper]{article}

\usepackage{iccv}
\usepackage{times}
\usepackage{epsfig}
\usepackage{graphicx}
\usepackage{amsmath}
\usepackage{amssymb}
\usepackage{graphicx}
\usepackage{amsmath}
\usepackage{amssymb}
\usepackage{booktabs}

\usepackage[utf8]{inputenc} 
\usepackage[T1]{fontenc}    
\usepackage{url}            
\usepackage{booktabs}       
\usepackage{amsfonts}       
\usepackage{nicefrac}       
\usepackage{microtype}      
\usepackage{xcolor}         
\usepackage{graphicx}
\usepackage{wrapfig}
\usepackage{tabularx}
\usepackage{multirow}
\usepackage{amsmath}
\usepackage{mathrsfs}
\usepackage{makecell}
\usepackage{booktabs}

\usepackage[breaklinks=true,letterpaper=true,colorlinks,bookmarks=false]{hyperref}
 \iccvfinalcopy 

\def\iccvPaperID{8712} 

\begin{document}

\title{DeLR: Active Learning for Detection with Decoupled Localization and Recognition Query}

\author{Yuhang Zhang$^1$, Yuang Deng$^2$, Xiaopeng Zhang$^1$, Jie Li$^2$, Robert C. Qiu$^3$, Qi Tian$^1$
\\ $^1$Huawei Inc.~~$^2$SJTU~~$^3$HUST
\\ {\tt\small \{lijiecs, rcqiu\}@sjtu.edu.cn, tian.qi1@huawei.com
}
\\ {\tt\small \{cupcake3419, zxphistory\}@gmail.com, lnsydengyuang@163.com
}
}
\maketitle

\begin{abstract}
Active learning has been demonstrated effective to reduce labeling cost, while most progress has been designed for image recognition, there still lacks instance-level active learning for object detection. In this paper, we rethink two key components, \emph{i.e.,} localization and recognition, for object detection, and find that the correctness of them are highly related, therefore, it is not necessary to annotate both boxes and classes if we are given pseudo annotations provided with the trained model. Motivated by this, we propose an efficient query strategy, termed as \textbf{DeLR}, that \textbf{D}ecoupling the \textbf{L}ocalization and \textbf{R}ecognition for active query. In this way, we are probably free of class annotations when the localization is correct, and able to assign the labeling budget for more informative samples. There are two main differences in DeLR: 1) Unlike previous methods mostly focus on image-level annotations, where the queried samples are selected and exhausted annotated. In DeLR, the query is based on region-level, and we only annotate the object region that is queried; 2) Instead of directly providing both localization and recognition annotations, we separately query the two components, and thus reduce the recognition budget with the pseudo class labels provided by the model. Experiments on several benchmarks demonstrate its superiority. We hope our proposed query strategy would shed light on researches in active learning in object detection.
\end{abstract}

\section{Introduction}
\label{sec:intro}

As a basic computer vision task, object detection has experienced remarkable progress over the past decade. However, training a high performance object detector usually requires a large scale of images with detailed bounding box annotations, which is time-consuming and sometimes redundant. There are several alternatives to reduce the cost for bounding box annotations, an efficient way is to rely on weak supervision, where the bounding box annotations can be relaxed to image-level supervision \cite{Bilen2016Weakly,CMIL2019,MinEntropy2019} or click supervision \cite{Papadopoulos2017TrainingOC,Papadopoulos2017ExtremeCF}. On the other hand, semi-supervised learning merely makes use of data mining for annotations, which is efficient but suffers noisy labels. As a compromise, active learning, which can be treated as an advancement for simple semi-supervised learning, actively select samples for annotations, and can achieve competitive performance with limited labeling cost. However, active learning has been mostly designed for image classification~\cite{Nguyen2004active, Sener2018Active,LearningLoss19,huang2021Temporal}, while few works focus on object detection, since it should consider both localization accuracy and recognition correctness, and is a more complex decision procedure. For this problem, an oracle method is to provide accurate bounding boxes as well as class labels for image that needs to be annotated. We notice that different methods vary in sample selection strategy \cite{Choi_2021_ICCV,LocalAware18,PedesDetect19}, \emph{i.e.,} which sample to be annotated, while few works target at designing efficient query method, which we consider is important for active learning in object detection. 

\begin{figure*}
    \centering
   \includegraphics[width=\textwidth]{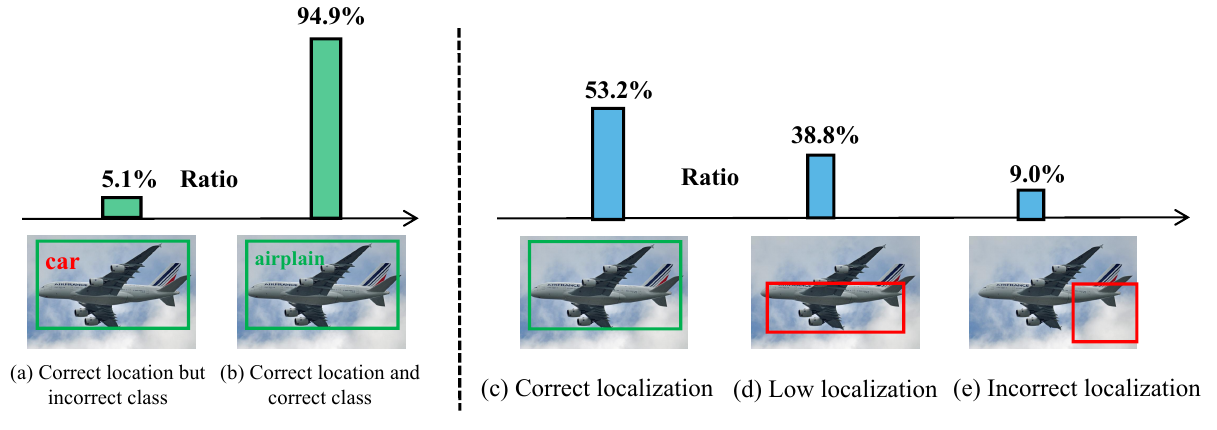}
   \caption{ Analysis of pseudo annotations on PASCAL VOC. Left: the relation between correct localization and recognition. Right: Summary of distribution for different localization cases. We find that the recognition is probably correct once the localization is solved, and thus motivates our decoupled query framework.}
   \label{fig:fisrt-page}
   \vspace{-0.35cm}
\end{figure*}

Considering the high cost of simultaneously offering box and class annotations, this paper rethinks active query strategy in object detection, and rise a question \emph{do we really need full annotations to samples that already endowed with pseudo annotations provided with models?} To answer this, we conduct a simple experiment, where we train a standard faster-RCNN detector on PASCAL VOC with randomly sampled $5\%$ images, and based on the detector, we assign pseudo annotations over the rest $95\%$ images as semi-supervised object detection (SSOD) does. To ease the noisy influence, we filter the boxes with a certain threshold (heuristically set as 0.7), and analyze the distribution of pseudo annotations in different cases. The localization can be categorized into three cases, \emph{i.e.,} (c) Correct localization (for correct, we define it as IoU between pseudo box and ground truth greater than a threshold, here simply set as 0.7); (d) Low localization accuracy (0.3<IoU<0.7); and (e) Incorrect localization (IoU<0.3). We find that even after box filtering, the correct location (IoU>0.7) is merely $53.2\%$, and most boxes suffer from low localization accuracy, which is harmful for retraining the detector if regarding them as ground truth. On the other hand, we further analyze the recognition results if the localization is correct, shown in the left part of Fig.~\ref{fig:fisrt-page}, the category accuracy achieves a high level (94.9\%) when the localization is correct. The above results reveal that in most cases, we only need to obtain accurate location information.

Based on the above analysis, we propose a novel active learning framework for object detection. Our method, termed as DeLR, which targets at \textbf{De}coupling \textbf{L}ocalization and \textbf{R}ecognition for active query. Such that the annotation cost can also be decoupled and those redundant annotations especially for label verification can be avoided for efficiency. In DeLR, we query the correctness of decoupled pseudo annotations to improve the utilization of the label budget. There are two main differences between DeLR with conventional active learning. First, previous methods mostly focus on image-level sample selection, where the queried samples are exhausted annotated, \emph{i.e.,} all objects are annotated within an image. While in DeLR, the query is based on instance-level, and we only annotate the object region that is queried. Second, the annotation $\mathcal{A}$ in the detection task including bounding box coordinates $\mathcal{A}_{loc}$ as well as class label $\mathcal{A}_{cls}$. In DeLR, we decouple the two types of annotations for query. For example, we provide box annotations for those regions with high location uncertainty, and keep using its pseudo class labels if we believe. Experiments demonstrate that such an effective annotation form can greatly improve the utilization under a fixed labeling budget. 

We summarize the contributions as follows:

$\bullet$ We present a novel active learning framework for object detection, which decouples the full annotations into location and category, and conduct an instance-level based decoupled query. In this way, we are able to avoid redundant annotations especially for category labels for efficient active learning. We hope that such a framework would shed light on the active learning for detection field and offer one possible direction for future research.

$\bullet$ We conduct detailed comparison experiments from multiple perspectives to demonstrate the effectiveness of DeLR, and achieve state-of-the-art performance in several widely used benchmarks under the same labeling budget. 

\begin{figure*}
    \centering
    \includegraphics[width=\textwidth]{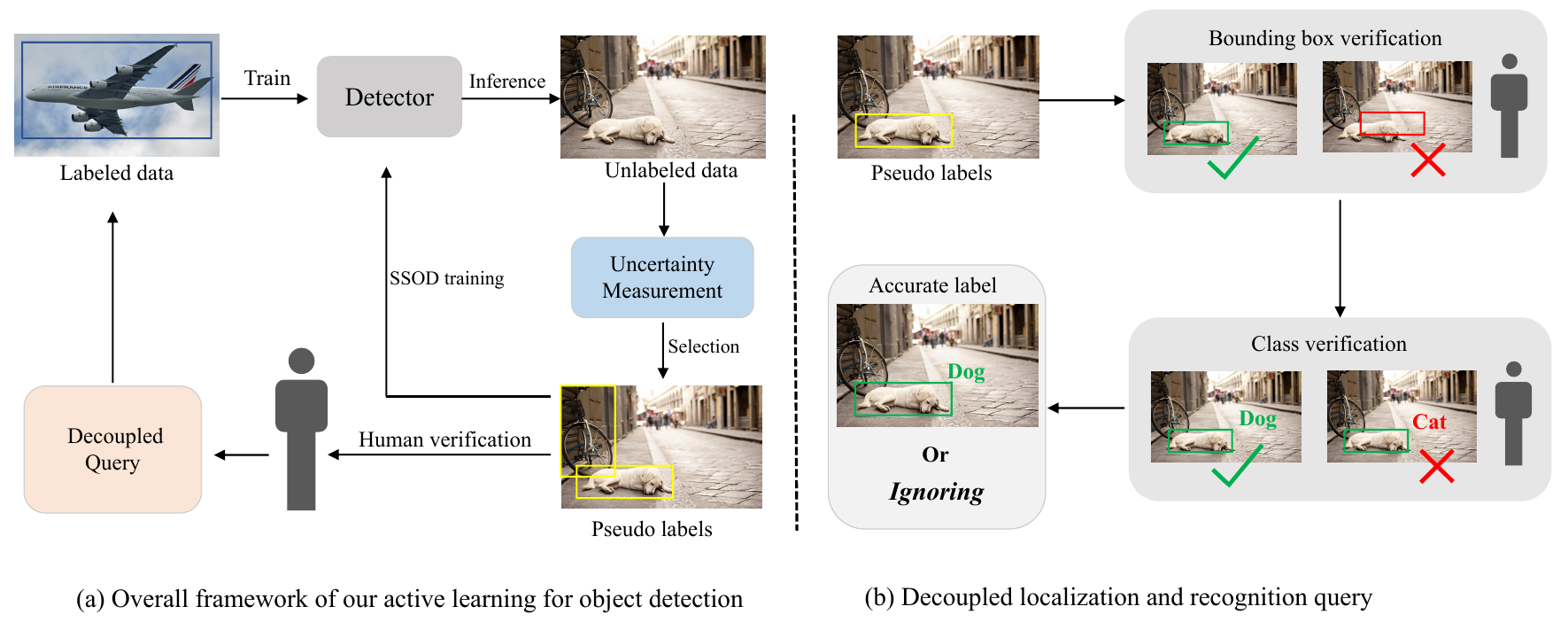}
    \caption{An illustration of our pipeline. (a) We first train the detector with initial labeled dataset, then leverage it to generate pseudo annotations for unlabeled data. After that, we evaluate the decoupled uncertainty of each pseudo annotation in terms of localization and recognition. (b) Detailed verification procedure, which needs the annotator to decide whether update the pseudo annotation, keep it or drop it.}
    \label{fig:pipeline}
\end{figure*}

\section{Related Work}
\paragraph{Active Learning for Classification and Detection.} Active learning aims to minimize the labeling cost by selecting the most valuable data that can improve the model performance\cite{NIPS2007_8f855179,Dasgupta2008hierachical,CDAL20,CoreSet18}, and plays an important role in modern machine learning systems. Most previous active learning methods focus on classification problem. 
According to the criterion of hard example selection, these works can be categorized into two aspects. The first is uncertainty-based methods that actively select the most uncertainty samples during training, which usually leverage the probability distribution of the predicted class \cite{Lewis1994sequential}, or the max entropy of posterior probabilities \cite{Wang_2017} and margin between the probability of the first and the second prediction \cite{MultiClass09,MarginBased06}. The second is the diversity-based approach, which selects diverse samples by estimating the whole distribution of the unlabeled pool \cite{7410873,6909473,Nguyen2004active}. Among them, Core-set \cite{Sener2018Active} is a typical diversity-aware method based on the core-set distance of intermediate features. 

With the development of research on active learning, a more difficult problem, active learning for object detection has attracted much attention recently. The detection task suffers more complex characteristics and is more challenging than classification. Some works specified for classification task are directly applied to the detection tasks, like \cite{LearningLoss19}, which simply sorting the loss prediction of instance in an image. \cite{Yuan_2021_CVPR} models the relationship between the instance uncertainty and image uncertainty for informative image selection. \cite{PedesDetect19} obtains the image-level uncertainty by calculating the uncertainty of background pixels. 
\cite{LocalAware18} focuses on the location accuracy, and defines two different scores to measure the location uncertainty of samples from different angles. 
In particular, a different solution is given by \cite{Choi_2021_ICCV}, which combines the detection pipeline with the Gaussian mixture model (GMM), and can directly obtain the localization and classification uncertainty after training. However, most previous methods provide full annotations for all objects in uncertain images, which will waste the label budget. In contrast, DeLR presents a region-level annotations and decouples the label information for more efficiently use.

\paragraph{Omni-supervised Learning.} Correct annotating detection dataset is expensive and time-consuming, which requires annotators to label not only category but also bounding box coordinate for all object in an image. For example, drawing one box needs 35 seconds \cite{Su2012CrowdsourcingAF}, and average annotation time of annotating per COCO image with categories is 346 seconds \cite{Ren2020UFO2AU}. To alleviate this problem, many methods focus on learning with few labels or combining with different forms of annotations to improve performance. Semi-supervised object detection \cite{ConsistencySelf20, Xu2021EndtoEndSO,Yang2021InteractiveSW,Tang2021HumbleTT} aims at training the detector with few labeled data, which is usually based on pseudo labels and mean-teacher pipeline to build the model. \cite{Papadopoulos2016WeDN} conducts a verification operation to improve detection, \cite{Papadopoulos2017TrainingOC} proposes a pipeline to train with click supervision. \cite{Ren2020UFO2AU} first proposes a unified framework to combine different types of annotations to train the detector, such as point, tags, box, scribbles, and so on. \cite{Wang2022OmniDETROO} takes a step in this direction and demonstrates that a mixture of annotations is better than full annotations when given a fixed labeling budget. Similar conclusion is also given in \cite{Desai2019AnAS}. 
In comparison, we incorporate the mixed annotations in active learning, and provide decoupled supervision according to its corresponding uncertainty.

\section{Method}
In this section, we first give an overview of the proposed DeLR framework for active learning in object detection. Then we elaborate the query procedure that decouples localization and recognition. At last, we present the sample selection strategy in our active learning pipeline.

\begin{figure*}
    \centering
   \includegraphics[width=\textwidth]{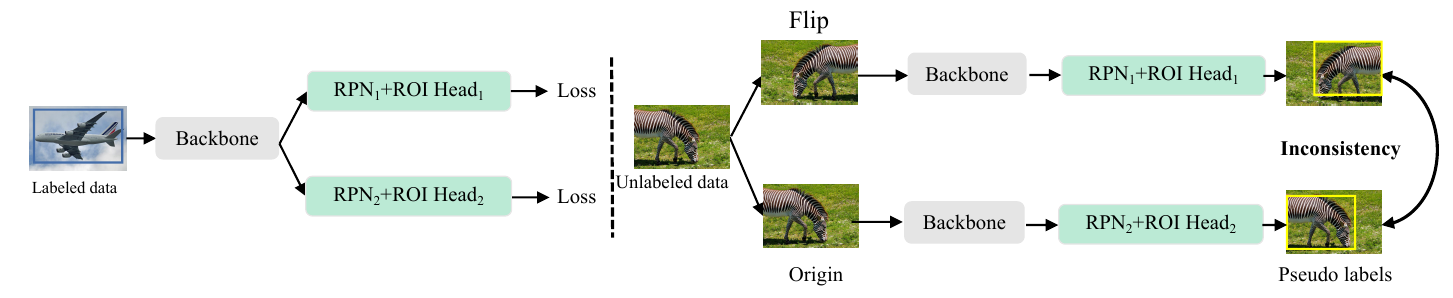}
   \caption{An illustration of our uncertainty calculation method. To enhance diversity, we train two RPN and ROI heads from random initial with the same input. These two different will be used to evaluate the inconsistency of two different views on the same unlabeled sample.}
   \label{fig:uncertainty}
\end{figure*}
\vspace{0.08in}

\subsection{Overview}
The overall framework of DeLR is illustrated in Fig.~\ref{fig:pipeline}(a). 
Given an unlabeled sample set, we first generate pseudo annotations for each unlabeled sample, by using the detector trained with initial labeled set. Then we evaluate the uncertainty of each pseudo annotation, using a simple consistency metric, which would be detailed in Sec. \ref{Consis_chapt}. Different from previous works that measure the uncertainty at image-level, we consider the annotation quality at fine-grained instance level, and most importantly, we rethink the annotations via decoupling boxes and classes, and measure both location and classification uncertainty for each pseudo annotation. 
The corresponding query are also decoupled, according to the priority given by the uncertainty metric, we decide whether to ask the annotator for boxes or classes verification, respectively. We claim that such decoupled query procedure are highly efficient than traditional ways, and is able to save annotation time substantially. 



\textbf{Active Learning Setup. } To begin with, we first formally define the active learning settings.
For conventional pool based active learning, we usually have a small set of labeled data $\mathcal{D}^l = \{x^l_i, y^l_i\}^{N_l}_{i=1}$ and a deep detection model $\mathcal{M}$ (\emph{e.g.}, Faster RCNN) to initialize the detector. Given the annotation budget $\mathcal{B}$, we decouple the annotation into location cost $\mathcal{B}_{loc}$ and recognition cost $\mathcal{B}_{cls}$. Then we can use the detector to decide which component of the pseudo annotation in $\mathcal{D}^u = \{x^u_i \}^{N_u}_{i=1}$ needs to be queried according to the designed sample selection strategy, and feed it to the labeled pool $\mathcal{D}^l$ according to the query result. Finally, we can leverage the updated $\mathcal{D}^l$ combined with unlabeled data (provided with pseudo annotations by the model) to retrain the detector, and these training and selecting procedures repeat until the annotation budget is used up. During the above procedure, two components count for the final performance, \emph{i.e.,} the choice of samples for query and the detailed query strategy, which we would elaborate in detail in the following.

\subsection{Decoupling Localization and Recognition Query}

Specifically, for the unlabeled dataset of size $N$, we first obtain the pseudo label $\{(b_i,c_i), i=1,2...,K\}$ on them, where $K$ is the total number of pseudo labels and usually $K \gg N$, and these pseudo labels will be filtered by a confidence threshold $\tau$. 
Then we can get the location uncertainty $u^i_{loc}$ and classification uncertainty $u^i_{cls}$ of each pseudo annotation through Eq.(~\ref{eq:uncertainty}), detailed in Sec. \ref{Consis_chapt}. After that, we rank the pseudo annotations in descending order according to their uncertainty metric, respectively. Based on the uncertainty in localization and recognition, we elaborate the query procedure as follows.

\textbf{Boxes Verification. } 
As described in Fig.~\ref{fig:fisrt-page}, the localization problem is a more difficult and challenging issue compared with classification. Therefore, in DeLR, we first probe the localization accuracy of the pseudo boxes.
We start with the pseudo boxes with the highest ranking of location uncertainty $u_{loc}$, \textbf{regardless of its class labels}, and verify its location accuracy by calculating the IoU \footnote{Considering that the annotator may not accurately decide the IoU value, we add a random disturbance to simulate the annotation procedure, detailed in Experiments.} between pseudo box $b$ and ground truth $b_{gt}$. We send the pseudo boxes (in practice, we enlarge the pseudo boxes twice as the annotation regions) to the annotator, and divide the results into three cases:

$\bullet$ Case One: If $\textit{IoU} (b,b_{gt}) \geq \textit{IoU}_{pos}$, we consider this pseudo box is correct, The pseudo coordinates will be preserved, and the annotator return a yes verification of the boxes and does not need to provide extra location information for this object;

$\bullet$ Case Two: If $\textit{IoU} (b,b_{gt})<\textit{IoU}_{bg}$, we consider this pseudo box is false positive detection, this pseudo label will be discarded directly to avoid bringing noise to the model training. In this situation, the annotator need return a no verification for this object; 

$\bullet$ Case Three: if $\textit{IoU}_{bg}$<IoU($b,b_{gt}$)<$\textit{IoU}_{pos}$, we correct $b$ by leveraging the $b_{gt}$ which has the highest IoU with it, denoted as $b$ = argmax$_{b_{gt}}(\textit{IoU}(b,b_{gt}))$. In this situation, the annotator need to return a new box that most closely with the provided pseudo box. In rare cases, the annotator would not return any valid box within the offered region, and this attribute to Case 2; 

Where $\textit{IoU}_{pos}$, $\textit{IoU}_{bg}$ are heuristically set as 0.7 and 0.3, respectively.

\textbf{Classes Verification. } 
After the boxes verification, we continue to verify the accuracy of its category $c$ based on the localization query. According to the results shown in Fig.~\ref{fig:fisrt-page}, the classification precision will achieve a high level while the localization is accurate, so we divide this verification process into two cases to further improve the utilization of annotations.

$\bullet$ Case One: If max($c$)>0.9 and $u^i_{cls}$ < $\tau_{cls}$, where $\tau_{cls}$ is the median of all classification uncertainty score, we consider this pseudo category is correct and will be preserved. In this situation, the annotator does not return any information. 

$\bullet$ Case Two: If max($c$) or uncertainty metric $u^i_{cls}$ does not meet the conditions above, we would conduct classes verification. It is simply by observing whether the pseudo category $c$ is equal to the ground truth class $c_{gt}$. Similar to the operation in box verification, we will keep the category $c$ if it is correct and replace it with $c_{gt}$ while it is wrong.

\subsection{Consistency Based Sample Selection} \label{Consis_chapt}
In this section, we introduce how we select uncertain instances from unlabeled sample for active learning. In our approach, we simply denote the instances whose output are not robust from different views as uncertain. In particular, we define the model uncertainty metric via diagnosing the consistency of the output between different views of same input.

Creating diversity for input is an important part of measuring consistency. In DeLR, in addition to using different views of samples, we train the detector consisting of a shared backbone, but with different RPN and ROI heads, and these two modules are randomly initialized differently, the whole diagram is shown in Fig.~\ref{fig:uncertainty}. Specifically, we first train the detector with two different heads using initial labeled data. After that, given an unlabeled data $x$, we perform a horizontal flip to obtain its augmentation view $\hat{x}$. Then we feed these two views into the pretrained detector, and through different RPN and ROI heads to get pseudo annotations $\{(b,c)\}$ and $\{(\hat{b},\hat{c})\}$, where $b$ and $\hat{b}$ contains 4 box coordinates $\{(x,y,w,h)\}$. Finally, we can obtain the uncertainty of each pseudo annotation through the following functions:

\begin{equation}
\begin{aligned}
u_{loc} = \frac{1}{4} \sum^4_{k=1} ||b_k-\hat{b}_k||,~~~u_{cls} = \mathcal{H}(c||\hat{c}),
\end{aligned}
\label{eq:uncertainty}
\end{equation}

where $||\cdot||$ is L1-loss and $\mathcal{H}(\cdot||\cdot)$ represents KL divergence. Note that for a pseudo label $\{(b,c)\}$, we can obtain its location uncertainty $u_{loc}$ and classification uncertainty $u_{cls}$ respectively through Eq.(~\ref{eq:uncertainty}), and provide corresponding types of annotation for those objects with high uncertainty.

\section{Experiments}
In this section, we conduct extensive experiments on several benchmarks to validate the effectiveness of our proposed active learning framework, as well as detailed ablation studies with hyperparameters and different component analysis.

\begin{figure*}
    \centering
   \includegraphics[width=0.925\textwidth]{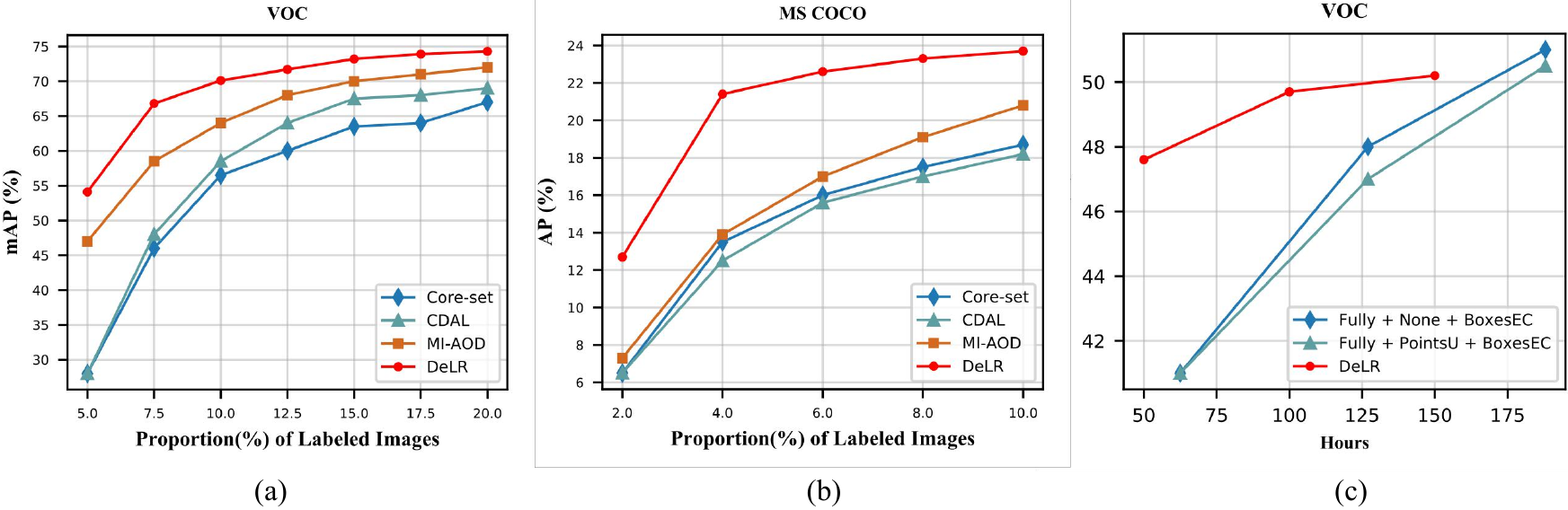}
   \caption{ Comparison with conventional active learning methods for object detection  on (a) VOC and (b) COCO. (c) Comparison with Omni-supervised object detection method on VOC.}
   \label{fig:main—results_1}
\end{figure*}

\begin{figure*}
    \centering
   \includegraphics[width=0.925\textwidth]{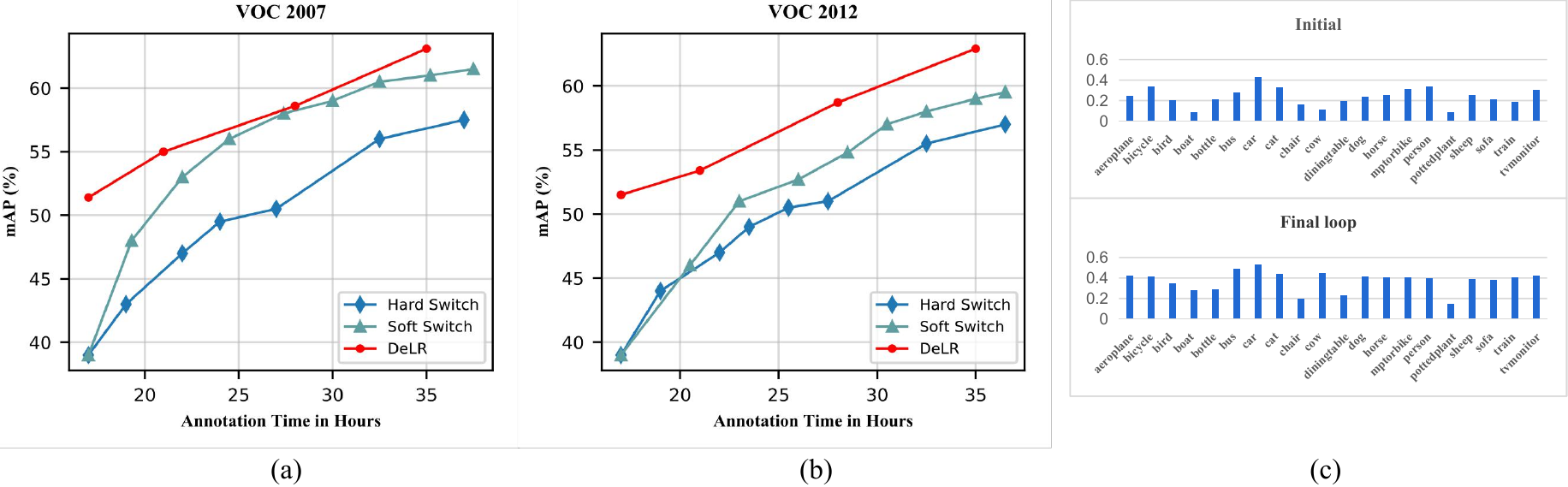}
   \caption{Comparison with adaptive active learning for object detection methods on (a) VOC 07 and (b) VOC 12. (c) The accuracy per class of DeLR in the zeroth and last cycle. }
   \label{fig:main—results_2}
\end{figure*}

\subsection{Experimental Settings}

\textbf{Datasets: } 
We evaluate DeLR on several standard detection benchmarks, including PASCAL VOC 2007/2012 \cite{Pascal10} and MS-COCO \cite{Lin2014MicrosoftCC}. PASCAL VOC includes 20 categories, the trainval set of VOC 2007 and 2012 consist of 5,011 and 11,540 images, respectively, while the VOC 2007 test set is used for testing for both datasets. MS COCO dataset is a large-scale challenging dataset that contains about 118k images for training ranging 80 classes. 

\textbf{Active Learning Settings: } 
In order to better measure the annotation cost of each type of supervision. In this paper,  we offer the annotation effort in terms of time, following the acknowledged metric of previous works. We summary the annotation time of each supervision as follows. It should be noted that unless specified, the time cost below refers to annotating one object, not the whole image.

\emph{\textbf{Localization and recognition verification.} } Verifying whether the location is correct takes around 1.6 second \cite{Papadopoulos2016WeDN}, while verifying category needs 2.7 second. \cite{Hangonebit22}. 

\emph{\textbf{Drawing a box.} } Drawing one rectangular box takes around 35 seconds according to \cite{Ren2020UFO2AU}

\emph{\textbf{Assigning category labels.} } We approximate this annotation time relying on the annotation and dataset statistics presented in previous work \cite{2009ImageNet,hu2020one,Hangonebit22}. Labeling an image in ImageNet \cite{2009ImageNet} need around 1 minute, according to the function used in \cite{hu2020one,Hangonebit22}, we consider
annotating a proposal with category takes 38 seconds for COCO and 26 seconds for PASCAL VOC, which is proportional the the total number of categories needs to be assigned. 

\emph{\textbf{Full annotation for an image.} } To make fair comparisons with previous full annotation methods, we need to convert the cost of annotating images to specific time. As reported in \cite{Wang2022OmniDETROO}, the average time of annotating all objects in one images is 346 seconds for COCO and 102.6 seconds for PASCAL VOC.

\textbf{Implementation Details:} 
We conduct all our experiment based on Faster RCNN \cite{FasterRCNN15} with ResNet-50 \cite{He2016Deep} and FPN \cite{FPN17}. Results based on other detector will be shown in the Appendix. For training with the initial label set, we set batch size as 2 per gpu, while for SSOD training after active learning verification the batch size is increase to 10 per gpu and 8 of them are unlabeled samples. The initial learning rate is set as 0.02 and would be decayed once by 0.1 during training. The weight decay and the momentum are set to 1e-4 and 0.9, respectively. As for other hyper-parameters, the threshold for filtering pseudo annotations $\tau$ is set to 0.7.
\subsection{Comparison with Conventional Active Learning Methods}
We first compare DeLR with several well known conventional active learning methods. including MI-AOD \cite{Yuan_2021_CVPR}, Core-set \cite{CoreSet18}, LL4AL \cite{yoo2019learning} and CDAL \cite{CDAL20}. 
For the VOC dataset, we randomly select 5\% of training data as the initial label set, and in each active learning, we have 2.5\% more images budgets from the rest unlabeled dataset until use up the annotation budget. As for the MS COCO, we use only 2\% labels to train the initial model, and select 2\% more labels in each active loop. In particular, for fair comparison, we convert the number of images that can be used in the labeled pool to the annotation time as our label budget. For example, 2.5\% labels in the VOC is equal to  16511*2.5\%*102.6/3600$\approx$11.8 hours to leverage . 
We trained the model with 6K iters for VOC and 18K iters for COCO.

\textbf{PASCAL VOC. } We first show the comparison results of PASCAL VOC in Fig.~\ref{fig:main—results_1}(a). It can be observed that DeLR outperforms all previous state-of-the-art methods with large margins at each active cycle. Specifically,  DeLR achieve about 7.1\%, 7.6\%, and 5.3\% advantages respectively when using 5.0\%, 7.5\%, and 10.0\% samples compared with MI-AOD. What's more, with 20\% labels, DeLR achieves 74.3\%, which surpasses MI-AOD by  about 2.0\%. These improvements demonstrate our progressive decoupling algorithm is a more efficient annotation method.

\textbf{COCO. } Then we evaluate DeLR on more challenging dataset MS COCO, which has more categories and more complex characteristics. It can be observed from Fig.~\ref{fig:main—results_1}(b) that DeLR also outperforms all compared methods at each active cycles, even with training iterations. Particularly, we surpass the previous state-of-art-method MI-AOD by 7.5\%, 5.8\%, 4.1\%  with 4\%, 6\%, and 8\% respectively. What's more, DeLR achieve 23.7\% with 10\% labels, which outperforms MI-AOD and Core-set by  about 2.9\% and 4.2\%. 

\begin{figure*}
    \centering
   \includegraphics[width=0.9\textwidth]{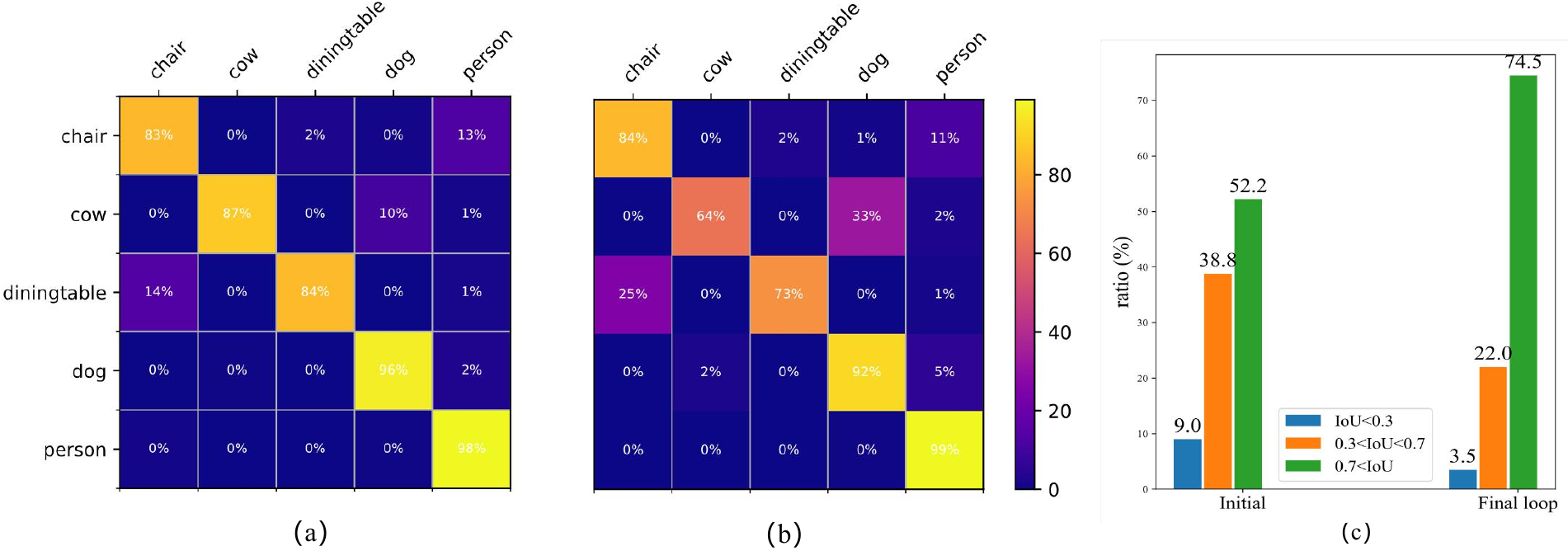}
   \caption{The confusion matrix of training with (a) DeLR and (b) SSOD. (c) Distribution of the accuracy of pseudo box in different active learning cycles.}
   \label{fig:con}
\end{figure*}

\subsection{Comparison with Adaptive Active Learning Methods}

In this section we compare DeLR with \cite{Desai2019AnAS}, which is also leverage different types of annotations to form a active learning for object detection pipeline. Different from DeLR, it uses full and click supervision \cite{Papadopoulos2017TrainingOC} to annotate uncertain samples. We make the comparisons on PASCAL VOC 2007 and PASCAL 2012 respectively, the initial model is trained with 500 images while the annotation budget is fixed to 35 hours. We train the model for only 4k iterations in each active cycle.

\textbf{VOC 2007. } The results of VOC 2007 are shown in Fig.~\ref{fig:main—results_2}(a), where the red and green line represents two different forms of \cite{Desai2019AnAS}. It can be observed that DeLR significantly outperforms \cite{Desai2019AnAS} in every active cycles, and DeLR can save more annotation time when achieving a similar performance. For example, we only need 21 annotation hours to achieve the mAP of 55\%, which saves about 3.6 hours and 9 hours respectively compared with soft switch and hard switch. while using the same annotation time of 30 hours, DeLR outperforms the previous forms hard switch by $\approx$ 6\%, and this trend is consistent with the increase of annotation hours, when the annotation time reaches 35 hours, DeLR surpasses the soft switch by 2\%. 

\textbf{VOC 2012. } The results of VOC 2012 are summarized in Fig.~\ref{fig:main—results_2}(b). Same as the results in VOC 2007, DeLR also outperforms \cite{Desai2019AnAS}in every active learning cycle. Specifically, to achieve the mAP of 55\%, DeLR cost about 23 hours annotation time, while soft switch and hard switch need 29 hours and 32.5 hours respectively. What's more, we improve the performance by 3.9\% compared with soft switch when use up the annotation budget.

\subsection{Comparison with Omni-supervised Learning Method}
In this section, to further demonstrate the superiority of DeLR, we compare it with Omni-DETR, a Omni-Supervised Object Detection (OSOD) method that combines different forms of annotations to improve detection \cite{Wang2022OmniDETROO}. Given a fixed annotation time budget of 150 hours, Omni-DETR trains the model with several different annotation policies, and aims to find the best one of them, here annotation policy refers to the strategy for mixing different annotation formats, including 1) Full+Unlabeled+Extreme Clicking Box (BoxesEC) \cite{Papadopoulos2017ExtremeCF}; 2) Full+Point \cite{Papadopoulos2017TrainingOC}+BoxesEC. The comparison experiment is conduct on the VOC dataset, and we train the model start from 10\% of all images for 16K iterations and use $\mathrm{AP}_{50:95}$ for evaluating.

\paragraph{Result. } The comparison results are summarized in Fig.~\ref{fig:main—results_1}(c). The green line and red line represent the above situation (1) and (2) respectively. It should be noticed that Omni-DETR conduct on the transformer based model Deformable DETR \cite{Zhu2021DeformableDD}, which has a stronger representation ability than our base model, even so, DeLR still outperforms it at every active cycle. Particularly, when given an annotation time of 100 hours, we surpass the Omni-DETR by 4.5\%, and we still have 1.0\% advantage when the annotation budget is fixed to 150 hours. 

\subsection{Analysis the Benefits of Pseudo Labels}

In this section we conduct the experiments to observe the improvement on the quality of pseudo labels with DeLR. We focus on the problem of location accuracy of pseudo label, which has been described in Fig.~\ref{fig:fisrt-page}. We recount the the distribution of IoU between pseudo box and ground truth after final active learning loop, and compare it with its initial state, the results are summarized in Fig.~\ref{fig:con}(c). It can be observed that there is a great improvement (+22.3\%) in the proportion of correct localization (IoU>0.7) in pseudo box, while the the proportion of low localization (0.3<IoU<0.7) is reduced to 22.0\% and the incorrect detection (IoU<0.3) is reduced to only 3.5\% after active query, which demonstrate DeLR can greatly improve the quality of pseudo labels.

We also observe in the AP of each category before and after decoupled query, and the results are summarized in Fig.~\ref{fig:main—results_2}(c). We found that the improvement of DeLR is particularly obvious for those difficult samples with poor initial performance. For example, the category \emph{cow} obtain only 10.8\% after initial training, and achieve 44.7\% after decoupled query and training, which obtain 33.9\% improvement and much greater than the average improvement of all categories 13.46\%. 

\begin{table}[]
\centering
 \small
\caption{The influence of different $\tau$  }
\vspace{0.08in}
\centering
\renewcommand\arraystretch{1.18}
\setlength{\tabcolsep}{2.0mm}
\begin{tabular}{c|c|c|c|c}
\Xhline{1.2pt}
$\tau$ & 5\% & 7.5\% & 10\% & 12.5\% \\
\hline
0.5 & 0.555 & 0.576& 0.610 & 0.651 \\
0.7 & 0.561 & 0.573 & 0.621 & 0.661 \\
0.9 & 0.543 & 0.569& 0.595 & 0.598 \\
\Xhline{1.2pt} 
\end{tabular}
\label{tab:thr_pl}
\end{table}

\begin{table}[]
 \small
\caption{The comparison of mAP between SSOD and DeLR based on random selection.}
 \vspace{0.08in}
\centering
\renewcommand\arraystretch{1.18}
\setlength{\tabcolsep}{1.0mm}
\begin{tabular}{c|c|c|c|c}
\Xhline{1.2pt}
Method & 5\% & 7.5\% & 10\% & 12.5\% \\
\hline
SSOD & 0.531 & 0.540 & 0.554 & 0.583  \\
 DeLR  &  0.552 & 0.594 & 0.626 & 0.670 \\
\Xhline{1.2pt} 
\end{tabular}
\label{tab:ann_random}
\end{table}

\subsection{Ablation Studies}

In this section, we conduct detailed ablation studies to analyze the advantage of our method. For efficiency, all experiments are conducted on PASCAL VOC dataset and trained with 3K iterations. 

\textbf{The Advantage of Decoupling Annotation.}
In this section we will demonstrate the advantage of our design decoupled annotation method. For fair comparison, we conduct DeLR based on randomly selecting pseudo labels to query, and compare it with the SSOD method, which trains the model on full annotations. This experiment aims to demonstrate that when given a fixed label budget, annotating an image with verification based decoupled supervision is better than full annotation. The results are summarized in Table~\ref{tab:ann_random}, it can be observed that DeLR still significantly outperforms SSOD methods at every active cycle.  What's more, we draw their confusion matrix and visualize random five categories in Fig.~\ref{fig:con}. From the results, it is obvious to see that the performance of decoupled annotation based method on the objects with the same category is also significantly better than the SSOD methods. The Above results demonstrate the effectiveness of our decoupled annotation method. 

\begin{table}[]
\centering
\small
\caption{The influence of different $\Delta_{pos}$. }
\centering
\vspace{0.08in}
\renewcommand\arraystretch{1.18}
\setlength{\tabcolsep}{2.0mm}
\begin{tabular}{c|c|c|c|c}
\Xhline{1.2pt}
\label{tab:delta_pos}
$\Delta_{pos}$ & 5\% & 7.5\% & 10\% & 12.5\% \\
\hline
0.05 & 0.565 & 0.580& 0.636 & 0.659 \\
0.10 & 0.580 & 0.592 & 0.632 & 0.658 \\
0.15 & 0.562 & 0.586 & 0.622 & 0.654 \\
0.20 & 0.569 & 0.589 & 0.620 & 0.647 \\
\Xhline{1.2pt} 
\end{tabular}
\end{table}
\begin{table}[]
\small
\caption{The influence of different $\Delta_{bg}$.}
\vspace{0.08in}
\centering
\renewcommand\arraystretch{1.18}
\setlength{\tabcolsep}{2.0mm}
\begin{tabular}{c|c|c|c|c}
\Xhline{1.2pt}
$\Delta_{bg}$ & 5\% & 7.5\% & 10\% & 12.5\% \\
\hline
0.05 & 0.563 & 0.575& 0.623 & 0.657 \\
0.10 & 0.557 & 0.596 & 0.635 & 0.662 \\
0.15 & 0.560 & 0.586 & 0.623 & 0.656 \\
0.20 & 0.569 & 0.581 & 0.640 & 0.659 \\
\Xhline{1.2pt} 
\end{tabular}
\label{tab:delta_neg}
\end{table}

\begin{table}
\centering
\small
\caption{The comparisons based on RetinaNet.}
\centering
\vspace{0.08in}
\renewcommand\arraystretch{1.18}
\setlength{\tabcolsep}{2.0mm}
\begin{tabular}{c|c|c}
\Xhline{1.2pt}
Labels & 5\% & 7.5\%  \\
\hline
Core-set & 29.1\% (-21.5\%)	&45.3\% (-15.4\%) \\
CDAL & 29.1\% (-21.5\%) & 47.9\% (-12.8\%) \\
MI-AOD & 47.2\% (-3.4\%) & 58.0\% (-2.7\%)\\
DeLR &	\textbf{50.6\%}	& \textbf{60.7\%} \\
\Xhline{1.2pt} 
\end{tabular}
\label{tab:retinanet}
\end{table}

\textbf{Robustness on different detectors}
Our method is a general active learning framework that decouples the localization and recognition in detection for efficient query, and can be easily combined with other detection models. 
In order to further prove our statement, we use RetinaNet \cite{RetinaNet20}, which is a typical one-stage detection framework, to evaluate our method. We conduct the experiment on VOC07\&12 dataset, and other settings are kept the same with the compared baseline. Due to the limitation of time and GPUs, we do not carefully adjust the training parameters, but it can be observed from Table~\ref{tab:retinanet} that our method is still significantly better than the previous active learning method, which also proves the robustness of our model.

\textbf{Hyper-parameters Analysis. }
In this section, we analyze the impact of several hyper-parameters, including the different disturbance $\Delta$ on the IoU threshold in boxes verification $\textit{IoU}_{pos}$ and $\textit{IoU}_{bg}$, and the category score threshold of filtering pseudo labels $\tau$.

\textbf{The Robustness of Verification IoU. } In our active learning pipeline, we set a fixed IoU threshold $\textit{IoU}_{pos}=0.7$ and $\textit{IoU}_{bg}=0.3$ to verify the accuracy of pseudo box. However, In practice, the annotators cannot accurately judge whether the pseudo box meets the IoU condition only by observation. For this reason, we add a disturbance $\Delta$ on the threshold to simulate the real situation. Specifically, every time we perform the verification, we randomly select a noise number from [0, $\Delta$], then add it to $\textit{IoU}_{pos}$ and $\textit{IoU}_{bg}$ as a new threshold. We conduct the experiments on different value of $\Delta$ on $\textit{IoU}_{pos}$ and $\textit{IoU}_{bg}$ respectively, and the results shown in Table~\ref{tab:delta_pos} and \ref{tab:delta_neg} demonstrate the robustness of our method.

\textbf{The Confidence Threshold $\tau$.}  $\tau$ is a parameters that mainly affect the SSOD training manner, only those pseudo labels whose maximum classification score is greater than $\tau$ will be used for SSOD training. It can be used to control the quality of pseudo labels, while previous SSOD methods usually set it in 0.9. The results are summarized in Table~\ref{tab:thr_pl}. It is interesting to find that 0.9 achieves the worst performance, even 5.3\% lower than 0.5 which has never been used in the previous SSOD training methods. The main reason is that our verification mechanism can filter out a lot of noise caused by the low threshold, and a lower threshold will provide more candidate samples for the verification mechanism.

\section{Conclusions and Limitations}

This paper proposed a novel active learning framework for object detection. Our method, termed as DeLR, which decouples localization and recognition for active query. Comparing with conventional full annotations, the advantage is that we are able to avoid redundant annotations especially for category labels according to the decoupled query. Experiments conducted on several detection benchmarks demonstrate the effectiveness of our proposed framework.
Although promising, DeLR is a new learning framework and still suffers some limitations. For example, DeLR verifies the pseudo annotations and still need provide the correct ones in some situations, which is the most time consuming step. It is hopeful that we are able to only conduct \emph{yes or no verification} to further improve the efficiency \cite{he2020momentum,chen2020simple}. Furthermore, the proposed framework still suffer lots of missing detection  due to the box filtering procedure, which is another core and not well solved problem in active learning. These remains to be studied in the future research.

{\small
\bibliographystyle{ieee_fullname}
\bibliography{egbib}
}

\end{document}